\documentclass[10pt,twocolumn,letterpaper]{article}

\usepackage{iccv}
\usepackage{times}
\usepackage{epsfig}
\usepackage{graphicx}
\usepackage{amsmath}
\usepackage{amssymb}

\usepackage{mathtools}
\usepackage{amsmath}

\usepackage{amssymb}
\usepackage{amsfonts}
\usepackage{amsopn}
\usepackage{graphicx} 
\usepackage{textcomp}
\usepackage{xfrac}
\usepackage{bbm}
\usepackage{bm}
\usepackage{overpic}
\usepackage{wrapfig}
\usepackage[ruled,vlined,linesnumbered]{algorithm2e}
\usepackage{multirow}
\usepackage{array}

\usepackage{color}
\definecolor{green}{rgb}{0, 0.5, 0}
\definecolor{orange}{rgb}{0.8, 0.6, 0.2}
\definecolor{red}{rgb}{1.0, 0.0, 0.0}
\definecolor{teal}{rgb}{0.0, 0.4, 0.4}
\definecolor{purple}{rgb}{0.65,0,0.65}
\definecolor{saffron}{rgb}{0.95,0.75,0.2}
\definecolor{turquoise}{rgb}{0.0,0.5,0.5}
\definecolor{brown}{rgb}{0.5, 0.16, 0.16}

\usepackage{overpic}

\newlength\savedwidth

\definecolor{lightgray}{rgb}{0.6, 0.6, 0.6}

\newcommand{\Fig}[1]{Figure~\ref{fig:#1}}

\usepackage[normalem]{ulem}

\newcommand{\hidecomment}[1]{}

\usepackage{xspace}

\usepackage[pagebackref=true,breaklinks=true,letterpaper=true,colorlinks,bookmarks=false]{hyperref}

\iccvfinalcopy

\ificcvfinal\fi

\begin{document}

\title{SOCS: Semantically-aware Object Coordinate Space for Category-Level\\6D Object Pose Estimation under Large Shape Variations}

\author{Boyan Wan\footnotemark[1] \qquad Yifei Shi\thanks{Joint first authors} \qquad Kai Xu\thanks{Corresponding author} \\
National University of Defense Technology\\
{\tt\small wanboyan@163.com \{yifei.j.shi,   kevin.kai.xu\}@gmail.com}
}
\maketitle
\ificcvfinal\thispagestyle{empty}\fi

\begin{abstract}
Most learning-based approaches to category-level 6D pose estimation are design around normalized object coordinate space (NOCS). While being successful, NOCS-based methods become inaccurate and less robust when handling objects of a category containing significant intra-category shape variations. This is because the object coordinates induced by global and rigid alignment of objects are semantically incoherent, making the coordinate regression hard to learn and generalize. We propose Semantically-aware Object Coordinate Space (SOCS) built by warping-and-aligning the objects guided by a sparse set of keypoints with semantically meaningful correspondence. SOCS is semantically coherent: Any point on the surface of a object can be mapped to a semantically meaningful location in SOCS, allowing for accurate pose and size estimation under large shape variations. To learn effective coordinate regression to SOCS, we propose a novel multi-scale coordinate-based attention network. Evaluations demonstrate that our method is easy to train, well-generalizing for large intra-category shape variations and robust to inter-object occlusions.

\end{abstract}

\section{Introduction}
\label{sec:intro}

6D object pose estimation, i.e. determining the 3D rotation and translation of a object in the camera coordinate system, is an important computer vision task with a large body of literature~\cite{brachmann2014learning,wang2019densefusion,labbe2020cosypose,li2018deepim}.
Category-level object pose estimation attempts to solve the problem without relying on the exact CAD model of the target object~\cite{wang2019normalized}, which is hence more challenging than instance-level one.
Since the seminal work of Wang et al.~\cite{wang2019normalized}, most existing category-level works are based on a canonical representation of Normalized Object Coordinate Space (NOCS). Given an unseen object instance, they learn a neural network to map the perspective projection of the object to the NOCS of the corresponding category from which object pose can be estimated.

\begin{figure}[t] \centering
\begin{overpic}[width=1.0\linewidth,tics=10]{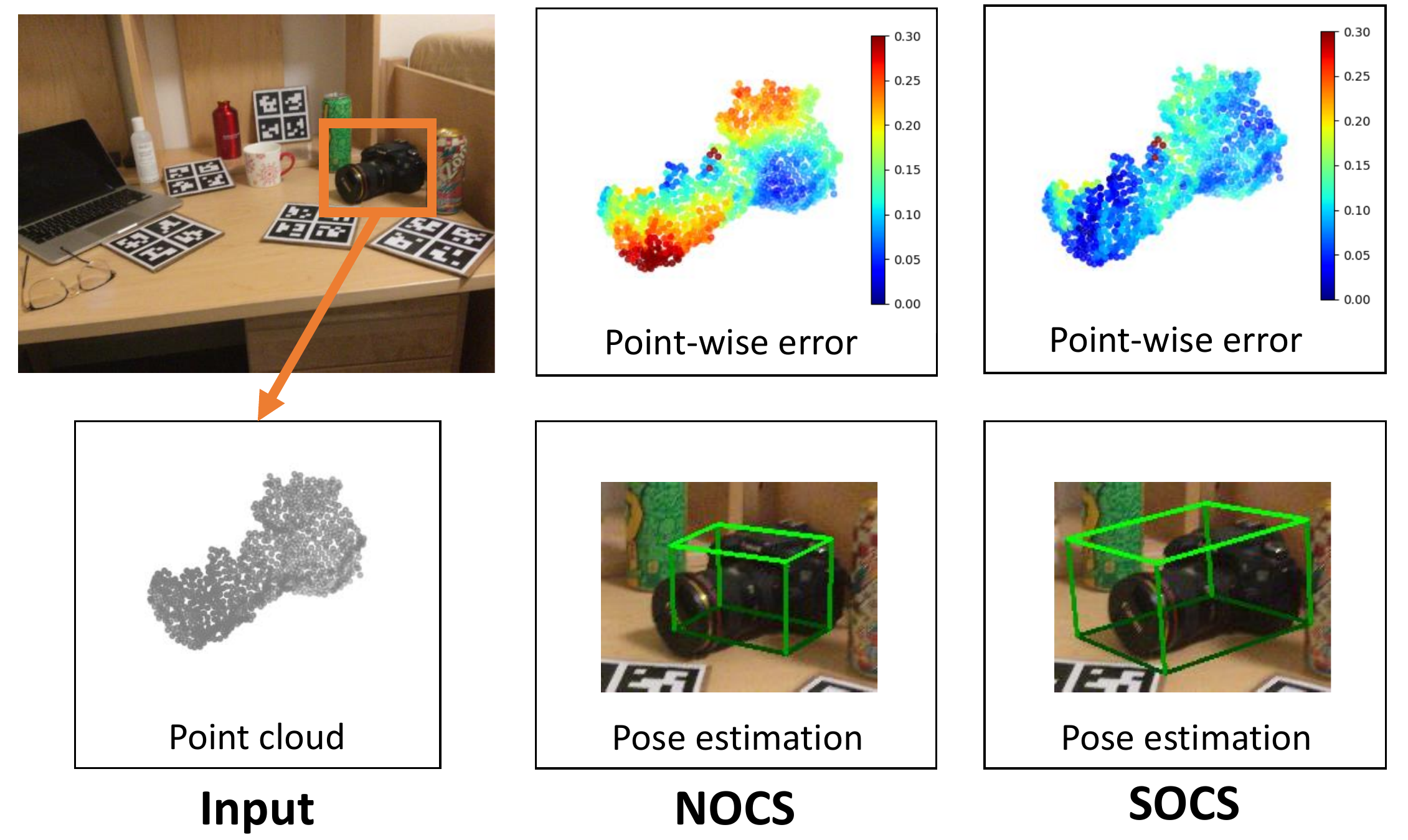}
   \end{overpic}
   \caption{NOCS~\cite{wang2019normalized}, constructed with globally aligned objects, finds difficulty in handling large intra-category shape variations. In this example, the coordinates regressed against NOCS for the long-lens camera contain much error (w.r.t. the CAD model under ground-truth pose and size) and the resulting pose is incorrect (middle). In contrast, the coordinates regressed against SOCS, built by semantically-guided non-rigid object alignment, are semantically coherent, leading to better pose estimation (right).}
   \label{fig:teaser}
\end{figure}

Given an object category, NOCS is defined by globally aligning a set of 3D object instances with normalized size and poses. It works well for objects with moderate intra-category shape variations.
When handling object categories containing significant shape variations,
however, NOCS-based methods become inaccurate and less robust.
This is because the object coordinates induced by global and rigid alignment are not semantically coherent. For instance, a point on the lens of a long-lens camera would be mapped to a semantically incorrect point in NOCS if the NOCS was constructed with camera models of significantly varying part proportions. Such misalignment makes the mapping network hard to learn and generalize, thus causing inferior pose accuracy under large shape variations (\Fig{teaser}).

To tackle this issue, we propose Semantically-aware Object Coordinate Space (SOCS) to achieve accurate and robust category-level 6D object pose and size estimation under large shape variations.
Unlike NOCS which is constructed by directly aligning pose and size normalized objects of a specific category, SOCS is built by \emph{warping-and-aligning the objects guided by a sparse set of keypoints with semantically meaningful correspondence}, leveraging the state-of-the-art category-specific keypoint selection and matching for a shape set~\cite{shi2021skeleton}. In particular, we align all objects of a specific category in the training set to the \emph{mean shape}~\cite{tian2020shape} of the set. We utilize 3D thin-plate spline warping~\cite{duchon1977splines} to ensure a smooth non-rigid deformation and hence coordinate interpolation. SOCS is therefore semantically coherent: Any point on the surface of a object can be mapped to a semantically meaningful location in SOCS, allowing for accurate pose and size estimation.


To learn the mapping from image space to SOCS effectively, we propose a novel multi-scale coordinate-based attention network. 
To capture the shape variation of the target object in image space, we devise a multi-scale feature extraction network with cross-attention feature aggregation.
In the cross-attention module, we encode global point positions to help better extract coordinate-sensitive features. Thanks to such global positional encoding, our network is able to model 3D points in the full space, which further enables a dense point sampling in SOCS training. The letter facilitates dense coordinate estimation even for unobserved locations, which is critical to handling inter-object occlusions.
To attain pose invariance, the network is trained in a contrastive fashion with a pose consistency loss.

We conducted extensive evaluations demonstrating that our method is
1) easy to train,
2) well-generalizing for large intra-category shape variations, and 3) robust to inter-object occlusions.
Even with the vanilla mapping network of~\cite{wang2019normalized}, our method is still comparable to state of the arts, clearly showing the effectiveness of SOCS.
Our full method achieves state-of-the-art on the NOCS-REAL275 and ModelNet40-partial datasets, improving the $5^{\circ}5$cm score by $5.6\%$ on NOCS-REAL275 and $5^{\circ}0.05$ score by $16\%$ on ModelNet40-partial.
In particular, ModelNet40-partial contains categories containing objects with large shape variations.


In summary, our work makes two contributions. \emph{First}, we propose semantically-aligned object coordinate space (SOCS) to accommodate large intra-category shape variations for semantically coherent coordinate regression. \emph{Second}, we propose a multi-scale attention network for learning the mapping from image space to SOCS effectively allowing for dense coordinate regression.

\section{Related Work}
\label{sec:related}

\subsection{Category-level pose estimation}
Category-level pose estimation aims to predict the pose of unseen instances from a single-view image without knowing their 3D model.
Existing work could be roughly classified into direct regression and correspondence-based methods.
Direct regression methods estimate object pose by extracting pose-sensitive features from the input~\cite{zhang2022rbp,lin2022category,deng2022icaps,liu2022catre,wang20206,wen2021bundletrack}.
The recent research focuses on exploring advanced network architectures~\cite{chen2020learning}, proper learning schemes~\cite{di2022gpv}, and different output representations~\cite{chen2021fs}.
Crucially, DualPoseNet~\cite{lin2021dualposenet} adopts two parallel pose decoders on top of a shared pose encoder, learning the consistency between the two brunches to impose complementary supervision.
FS-Net~\cite{chen2021fs} proposes a decoupled rotation output mechanism to complementarily estimate the rotation components.
Correspondence-based methods first estimate the correspondence between the observed points and its coordinate in the canonical space and then optimize pose and size by postprocessing. This requires methods to extract pose-invariant point features.
Wang et al.~\cite{wang2019normalized} present the representation of NOCS to enable the learning of pose for unseen objects.
Wen et al.~\cite{wen2022catgrasp} introduce NUNOCS, which allows non-uniform scaling across three dimensions, facilitating fine-grained dense correspondences across object instances with large shape variations.
Crucially, a bunch of recent works have adopted the categorical mean shape to facilitating the computation of correspondences between the observed points and their canonical coordinate~\cite{tian2020shape,chen2021sgpa,lin2022sar}.
Our method falls into the category of correspondence-based methods. However, it is different from previous work as it learns semantically-aware dense correspondences, resulting in more accurate results.

\subsection{Implicit field for pose estimation}
Many recent works have investigated implicitly representing 3D shapes with a continuous and differentiable implicit field implemented by neural networks.
While most of the research in this field focuses on shape reconstruction, a handful of methods adopt implicit fields to estimate object pose~\cite{peng2022self,agaram2022canonical}.
A straightforward way is to jointly reconstruct the object surface and estimate its pose~\cite{bruns2022sdfest,pavllo2022shape,li2022generative} with a unified framework.
For example, ShAPO~\cite{irshad2022shapo} jointly predicts object shape, pose, and size in a single-shot manner.
DISP6D~\cite{wen2022disp6d} disentangles the latent representation of shape and pose into two sub-spaces, improving the scalability and generality.
Neural Radiance Fields (NeRF)~\cite{mildenhall2021nerf} provides a mechanism for capturing complex 3D structures from only one or a few RGB images, which is also applicable to object pose estimation.
iNeRF~\cite{yen2021inerf} estimates pose for objects with complex geometry with a pre-trained NeRF model.
NeRF-Pose~\cite{li2022nerf} first reconstructs the object with NeRF and then estimates the object pose.
Unlike the traditional correspondence-based methods which predict 3D object coordinates at pixels of the input image, Huang et al.~\cite{huang2022neural} predict canonical coordinates at any sampled 3D in the camera frustum, generating continuous neural implicit fields of canonical coordinates for instance-level pose estimation.
Despite the similarity in the general concept, our method tackles the problem of category-level pose estimation where semantically-ware cross-instance correspondences need to be estimated.  

\section{Method}
\label{sec:method}

\paragraph{Overview.}
In this section, we first describe how to generate SOCS. Then, we present the multi-scale coordinate-based attention network for SOCS estimation. In particular, a surface-independent point sampling strategy and a pose-invariant feature extraction training scheme are introduced. Last, we elaborate on the details of network inference and pose estimation.

\subsection{SOCS}
\label{sec:seal_nocs}
Existing canonical coordinate spaces for category-level 6D pose estimation, such as NOCS~\cite{wang2019normalized}, are induced by global and rigid alignment, leading to semantic incoherency on the object coordinates.
When handling object categories containing significant shape variations, NOCS-based methods become inaccurate and less robust.
We introduce Semantically-aware Object Coordinate Space (SOCS) to alleviate the problem of NOCS.
The coordinates in SOCS are generated by the category-specific keypoints, allowing fine-grained non-rigid coordinate alignment.


Specifically, given the shapes $\{S_i\}$ of a category in the training set, we first generate the categorical mean shape $S_a$ by using the pre-learned autoencoder~\cite{tian2020shape}.
The coordinates of $S_a$ are regarded as the coordinates of SOCS.
To build correspondence between the object coordinate of any object instance $\{S_i\}$ and the SOCS, we detect the semantically consistent keypoints $\{K_i\}$ and $K_a$ for $\{S_i\}$ and $S_a$, respectively, by using the Skeleton Merger~\cite{shi2021skeleton}.
We denote the detected keypoints $K_i = \{k_j\}$, $j\in[1,m]$ and $K_a = \{k_j^{a}\}$, $j\in[1,m]$.
$m$ is the number of keypoints in a single shape.

Next, we compute the dense correspondence between $S_i$ and $S_a$ by considering the alignment of the semantically consistent keypoints. This is achieved by a 3D thin plate spline warping function~\cite{duchon1977splines}:
\begin{equation}\label{eq:sdf_ratio}
\begin{aligned}
\Phi(x)=c+b^{T} x+w^{T} &\mathbf{s}(x),\\
\mathbf{s}(x)=[\sigma\left(x-k_1^a\right), \sigma\left(x-k_2^a\right), \cdots, & \sigma\left(x-k_m^a\right)]^{{T}},\\
\sigma(x)=\|x\|_2^2 \cdot \log \|x&\|_2,
\end{aligned}
\end{equation}
where $c \in \mathbb{R}^{3}$, $b \in \mathbb{R}^{3 \times 3}$, $w \in \mathbb{R}^{m \times 3}$ are the parameters which are determined by optimizing the following function:
\begin{equation}
min \sum_{j=0}^{m}\left \|k_j^{a} - \Phi(k_j)\right \|^2.
\end{equation}
Once the parameters $c$, $b$, and $w$ are determined, for any coordinate $x$ in the object coordinate space, its SOCS could be computed as $x^{a}=\Phi(x)$.

\begin{figure}[t] \centering
\begin{overpic}[width=1.0\linewidth,tics=10]{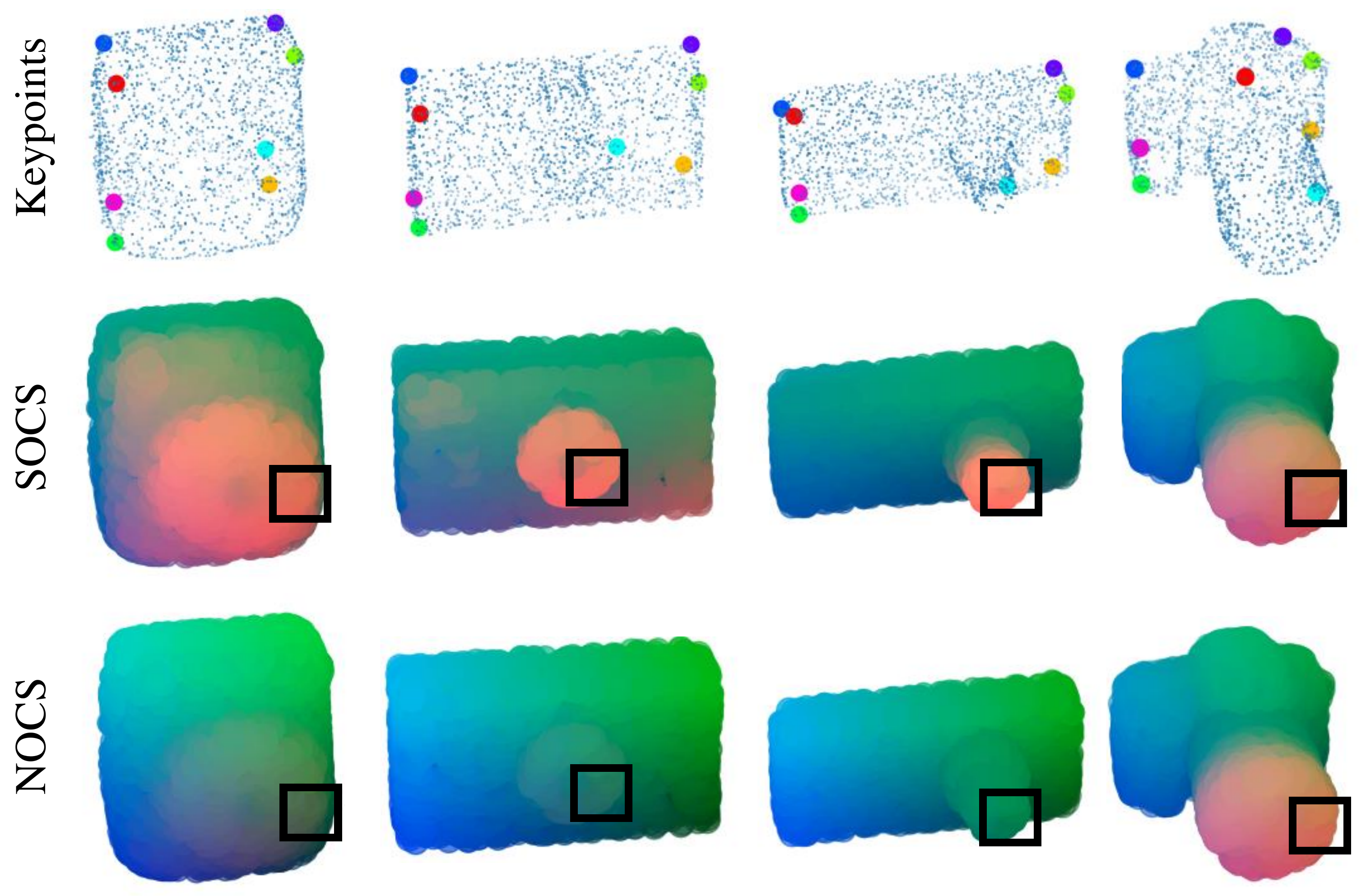}
   \end{overpic}
   \caption{The proposed SOCS is more semantically meaningful, facilitating the learning of correspondence for objects with large shape variations. The coordinates in the canonical space (represented by the color) of the same semantic part in different objects are similar in SOCS (Middle row) and dissimilar in NOCS (Bottom row). Please pay special attention to the highlighted regions.}
   \label{fig:heatmap}
\end{figure}

Compared to the NOCS representation and its variants~\cite{wen2022catgrasp} developed for category-level 6D pose and size estimation, the SOCS is more semantically meaningful, thus facilitating the learning of correspondence even for objects with large shape variations.
See \Fig{heatmap} for an illustration.

\subsection{Training of SOCS Estimation Network}
\begin{figure}[t] \centering
	\begin{overpic}[width=1.0\linewidth,tics=10]{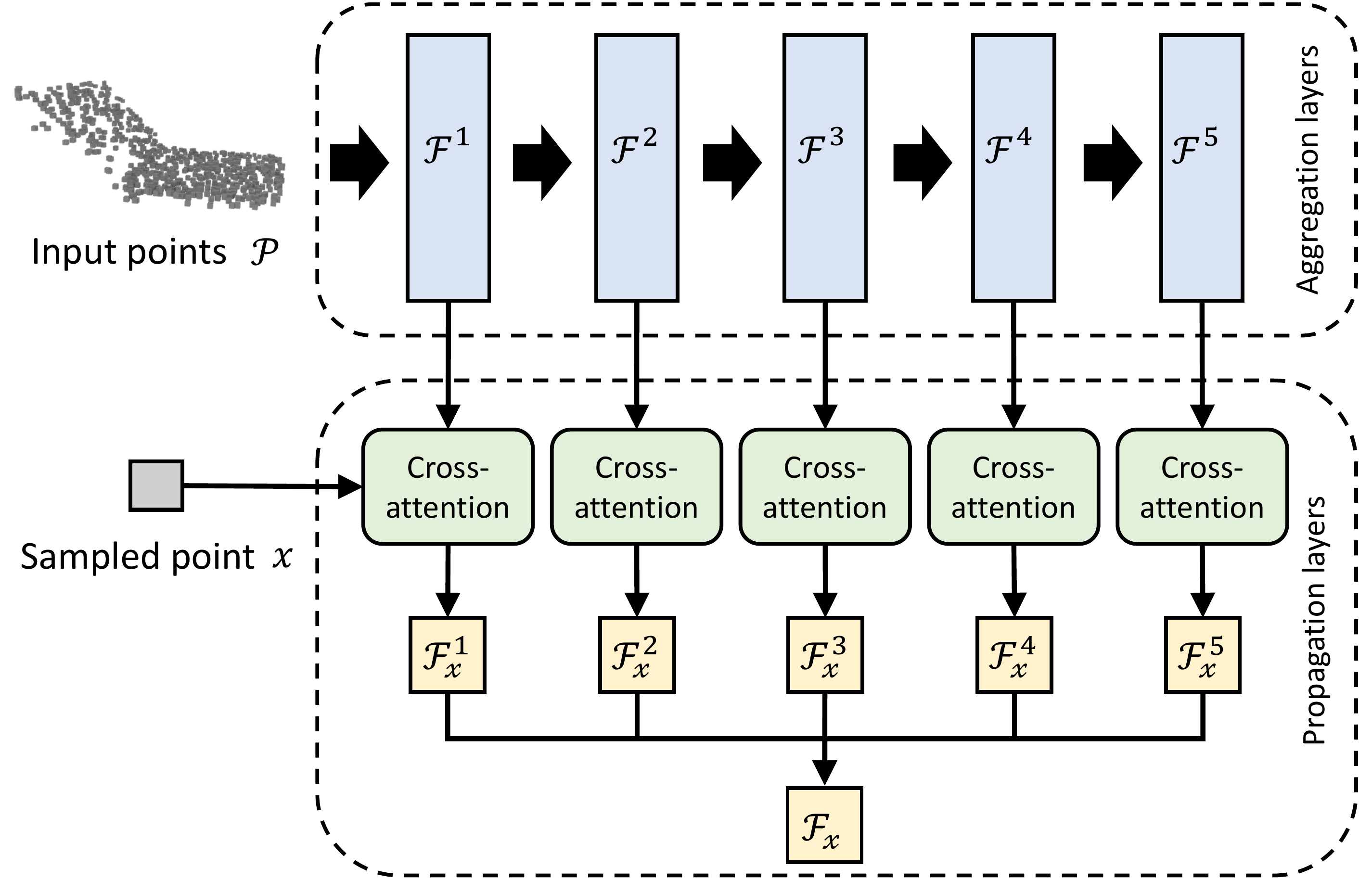}
   \end{overpic}
   \caption{Given an input point cloud $\mathcal{P}$, the aggregation layers generate the features $\{\mathcal{F}^\alpha$\}, $\alpha\in[1,5]$ at multiple network blocks. The propagation layers then propagate features from the input points to the sampled point $x$.
   }
   \label{fig:networks}
\end{figure}

In this section, we describe how to estimate the point-wise dense SOCS from an image.
Estimating SOCS from a single-view image is non-trivial due to the potential large shape variations and the inter-object occlusions.
To learn the mapping from input points to SOCS effectively, we propose a novel multi-scale coordinate-based attention network.
An overview of the network architecture is shown in \Fig{networks}.


\paragraph{Multi-scale coordinate-based attention network.}
The network contains two main components: \emph{aggregation layers} and \emph{propagation layers}.
The \emph{aggregation layers} extract per-point features from the point cloud.
The point cloud is cropped from the depth image of the detected object.
Since the task of category-level pose estimation could be challenging due to the large shape variations and severe occlusion of the input point cloud, we take 3D-GCN~\cite{lin2020convolution}, which is able to aggregate contextual information of 3D point clouds with good performance, as the backbone.
To be specific, the 3D-GCN takes the point cloud $\mathcal{P} \in \mathbb{R}^{n \times 3}$ as input, generates the downsampled points $\mathcal{P}^{\alpha}$ and extracts the features $\mathcal{F}^{\alpha}$ at the $\alpha$-th block, where $\alpha\in[1,5]$.
Note that the aggregation layers could be any other 3D point-based network backbone according to the practical requirements.
We found that 3D-GCN works best in our problem setting.

The \emph{propagation layers} are then developed to propagate the feature from the downsampled points $\{\mathcal{P}^{\alpha}\},\alpha\in[1,5]$ to any query point $x$ and estimate its SOCS $x^a$.
Extracting feature of unseen points is a non-trivial task, due to the infinite query location in the 3D space.
The ideal extracted feature should be both context-sensitive and coordinate-sensitive.
To achieve this, we propose an implicit neural network with coordinate-based multi-scale contextual feature propagations.

\begin{figure}[t] \centering
	\begin{overpic}[width=1.0\linewidth,tics=10]{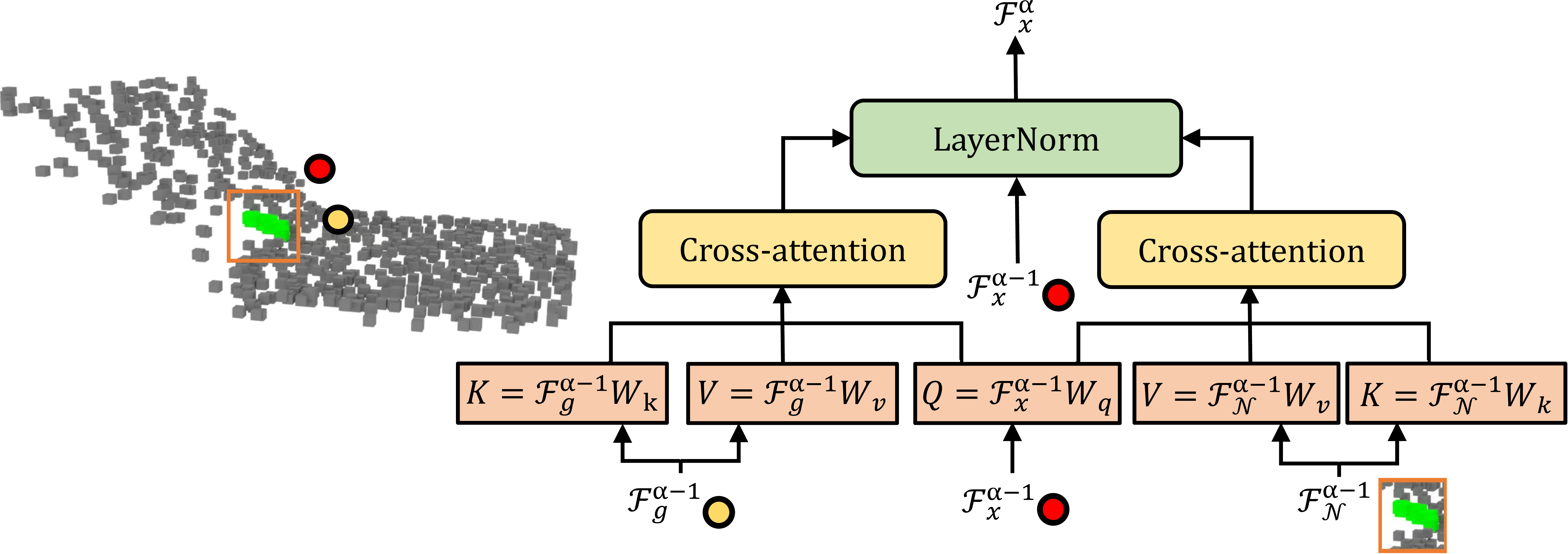}
   \end{overpic}
   \caption{The network architecture of the propagation layers. For a sampled point (gray), the query point (red) updates its feature by performing cross-attention operations with the k-nearest neighbors (green) and the global point (yellow), respectively.}
   \label{fig:attention}
\end{figure}

We first initialize the feature vector at query point $x$ as a zero vector, i.e. $\mathcal{F}_x^0=\bm{0}$, $\mathcal{F}_x^0 \in \mathbb{R}^{h}$, where $h$ is the feature length.
For each block, we update the feature with a cross-attention module (see \Fig{attention}).
Specifically, at the $\alpha$-th block, we compute the k-nearest neighbors $\mathcal{N}^{\alpha}\in \mathbb{R}^{16\times 3}$ of $x$ from $\mathcal{P}^{\alpha}$.
We denote the feature of $\mathcal{N}^{\alpha}$ as
$\mathcal{F}^{\alpha}_\mathcal{N}\in \mathbb{R}^{16 \times h}$.
Moreover, we introduce a global point $g^{\alpha}=\text{Mean}(\mathcal{P}^{\alpha})$ with the feature being $\mathcal{F}_g^{\alpha}=\text{Mean}(\mathcal{F}^{\alpha})$, where $\text{Mean}(\cdot)$ is the element-wise averaging operation.
The global positional encoding provides contextual information, enabling our network to extract features on 3D points in the full space. The letter facilitates dense coordinate estimation even for unobserved locations, which is critical to handling inter-object occlusions.

The update term on the feature $\mathcal{F}_x^{\alpha-1}$ is estimated by considering the relations to both $\mathcal{N}^{\alpha}$ and $g^{\alpha}$:
\begin{equation}\label{equ:cross_attention}
\begin{aligned}
\Delta^{\alpha}&=\text{Softmax}\left(\frac{\left(\mathcal{F}_\mathcal{N}^{\alpha} W_k\right)\left(\mathcal{F}_x^{\alpha-1} W_q\right)^{\mathrm{T}}+r}{\sqrt{\mathrm{h}}}\right)\mathcal{F}_\mathcal{N}^{\alpha} W_v\\
&+\text{Softmax}\left(\frac{\left(\mathcal{F}_g^{\alpha} W_k\right)\left(\mathcal{F}_x^{\alpha-1} W_q\right)^{\mathrm{T}}+r_g}{\sqrt{\mathrm{h}}}\right)\mathcal{F}_g^{\alpha} W_v
\end{aligned}
\end{equation}
where $W_q$, $W_k$, $W_v$ $\in \mathbb{R}^{h \times h}$ are the learnable weights. $r \in \mathbb{R}^{k}$ denotes the influence factor of $x$ to the points in $\mathcal{N}^{\alpha}$. Each element in $r$ is computed as follows by considering the relative position between the two points:
\begin{equation}\label{equ:element}
r_i=\text{EmbLayer}(x-\mathcal{N}^{\alpha}_i),
\end{equation}
where $\text{EmbLayer}(\cdot)$ is a two-layer MLP. Similarly, $r_g$ is computed to capture the position w.r.t. the center of input points:
\begin{equation}\label{equ:element}
r_g=\text{EmbLayer}(x-g^{\alpha}).
\end{equation}
Then, the updated feature at point $x$ is:
\begin{equation}\label{equ:addnorm}
\mathcal{F}_x^{\alpha} =\text{LayerNorm}(\mathcal{F}_x^{\alpha-1}+\Delta^{\alpha})
\end{equation}
where $\text{LayerNorm}(\cdot)$ is the layer normalization operation.

The extracted features at all the blocks are then concatenated: $\mathcal{F}_x=\text{Concat}(\mathcal{F}_x^1,\mathcal{F}_x^2,\mathcal{F}_x^3,\mathcal{F}_x^4,\mathcal{F}_x^5)$.
The concatenated feature is utilized to estimate the SOCS:
\begin{equation}\label{equ:estimation}
\begin{aligned}
    x^a_\text{X}=\text{Softmax}&(\text{MLP}_\text{X}(\mathcal{F}_x)),\\
    x^a_\text{Y}=\text{Softmax}&(\text{MLP}_\text{Y}(\mathcal{F}_x)),\\
    x^a_\text{Z}=\text{Softmax}&(\text{MLP}_\text{Z}(\mathcal{F}_x)),
\end{aligned}
\end{equation}
where $x^a_\text{X}$, $x^a_\text{Y}$, $x^a_\text{Z}$ are the predicted class denoting the coordinate in the axes of X, Y, Z, respectively.
$\text{MLP}_\text{X}(\cdot)$, $\text{MLP}_\text{Y}(\cdot)$, $\text{MLP}_\text{Z}(\cdot)$ represent multi-layer perceptrons.

Note that, our method is different from most existing methods where regression or classification with a small number of bins ($\text{B}<50$) for coordinate estimation are adopted.
We found that using a larger number of bins (e.g. $\text{B}=256$) in our method will not lead to the training being inefficient or failing to converge.
The advantage comes from the representation of SOCS, which greatly reduces the learning complexity.

\begin{figure}[t] \centering
	\begin{overpic}[width=1.0\linewidth,tics=10]{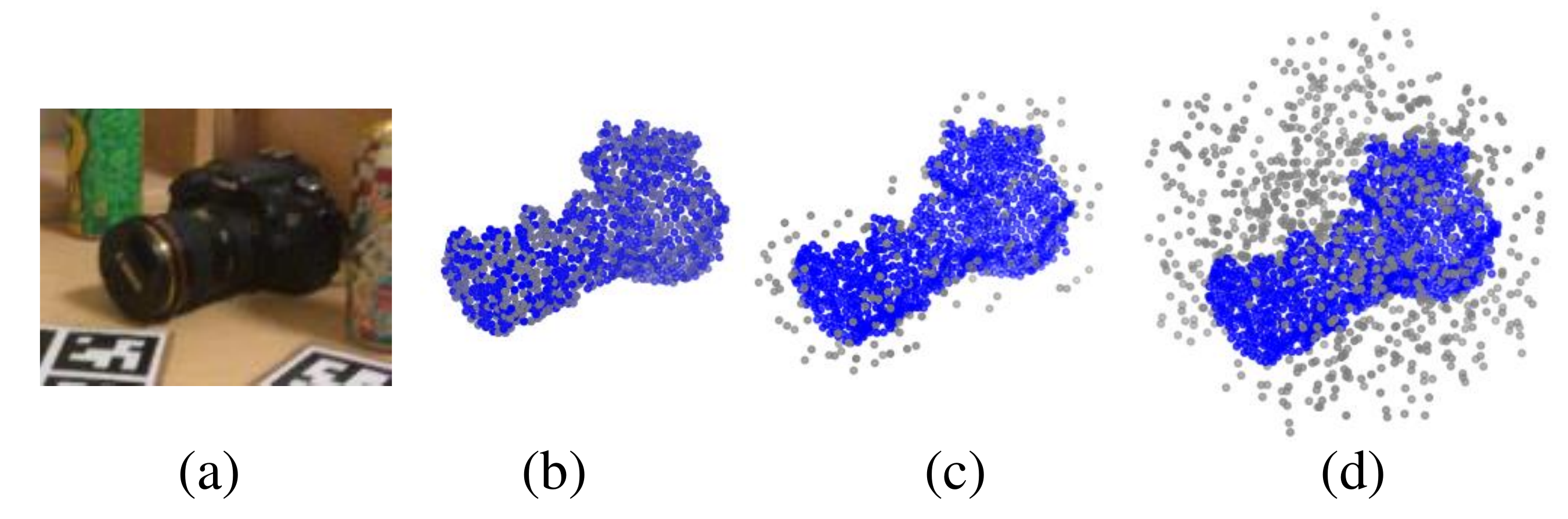}
   \end{overpic}
   \caption{Illustration of the point sampling strategies. (a) The input object. (b) Sampling from the input points. (c) Surface-dependent sampling. (d) Surface-independent sampling.}
   \label{fig:sampling}
\end{figure}

\paragraph{Surface-independent point sampling.}
We then describe how to sample points to feed into the multi-scale coordinate-based attention network.
Several sampling strategies could be considered (see \Fig{sampling}): 1) Sampling from the input points; 2) Surface-dependent sampling: random sampling near the input points; 3) Surface-independent sampling: random sampling in the whole 3D space.
We empirically found the surface-independent sampling strategy outperforms the others, thanks to the mechanism of global positional encoding.
Please refer to the result section for experimental analysis.
There are two reasons for this phenomenon.
First, sampling in the whole 3D space facilitates feature aggregation in the invisible region, bringing more global context information.
Second, sampling in the whole 3D space would decrease the overall pose estimation uncertainty, especially in scenarios where severe occlusion exists.

\paragraph{Network training.}
Next, we describe how to train the above network.
Suppose $\mathcal{X}$ is the set of sampled points, a naive loss function of the shape correspondence field estimation could be:
\begin{equation}\label{equ:loss_field}
\mathcal{L}_\text{SOCS}= \sum_{x\in \mathcal{X}} [\mathcal{L}_\text{CE}(x^a_\text{X},\hat{x}^a_\text{X})+\mathcal{L}_\text{CE}(x^a_\text{Y},\hat{x}^a_\text{Y})+\mathcal{L}_\text{CE}(x^a_\text{Z},\hat{x}^a_\text{Z})],
\end{equation}
where $\mathcal{L}_\text{CE}(\cdot)$ is the cross entropy loss, $\hat{x}^a_\text{X}$, $\hat{x}^a_\text{Y}$, $\hat{x}^a_\text{Z}$ denote the ground-truth.
However, we found that training the network is unstable and hard to converge, especially on categories with large shape variations.

To alleviate this issue, we adopt a contrastive training fashion with a pose consistency loss to further enhance the training.
The key insight is to learn the \emph{pose-invariant feature} by transforming the input point cloud, extracting its per-point features, and making the features of the initial point cloud and the transformed point cloud consistent.

Specifically, during training, we transform the input point cloud $\mathcal{P}$ with a random rigid transformation $\textbf{T}_\text{r}=\{\textbf{R}|\textbf{t}\}$.
We denote the transformed point cloud as $\mathcal{P}^{'}=\textbf{T}_\text{r} \cdot \mathcal{P}$.
Then, $\mathcal{P}$ and $\mathcal{P}^{'}$ are fed into the multi-scale coordinate-based attention network, respectively, to generate the per-point features.
For any point $x\in \mathcal{X}$ and the transformed point $x^{'}=\textbf{T}_\text{r} \cdot x$, the generated features should be consistent:
\begin{equation}\label{equ:loss_con}
\mathcal{L}_\text{consistency}=\sum_{x\in \mathcal{X}} ||\mathcal{F}_x-\mathcal{F}_{x^{'}}^{'}||_2,
\end{equation}
where $\mathcal{F}_x$ and $\mathcal{F}_{x^{'}}^{'}$ denote the extracted feature by the two network towers, respectively.
Overall, the training loss function is: $\mathcal{L}=w_\text{SOCS}\mathcal{L}_\text{SOCS}+w_\text{consistency}\mathcal{L}_\text{consistency}$, where $w_\text{SOCS}$ and $w_\text{consistency}$ are the pre-defined weights.


\paragraph{Training data preparation.}
To generate the training data of SOCS estimation, for each dataset, we first generate the dense SOCS for the complete 3D objects using the method described in Sec.~\ref{sec:seal_nocs}. The dense SOCS are then transformed into the camera coordinate with the 6D object pose.

\subsection{Network Inference, Pose and Size Estimation}
\paragraph{Network inference.}
During the inference, given an RGB-D image with untrained object instances in it, we first perform an object detection with Mask R-CNN~\cite{he2017mask}.
For each detected object, we crop the image, generate the point cloud, and feed them into the aggregation layers to generate features.
Then, we densely sample points in 3D space around the input points, and extract their features with the propagation layers, predicting the SOCS for every sampled point.

\paragraph{Pose and size estimation.}
The predicted per-point coordinate in SOCS is then transformed into the camera coordinate space with the transformation of the 6D object pose and a scaling operation.
The ideal transformation matrix $\textbf{T}\in \mathbb{R}^{4 \times 4}$ of the pose and the scaling matrix $\textbf{S}=\text{diag}(s_\text{X},s_\text{Y},s_\text{Z},1)$ should make the following function optimized:
\begin{equation}
min \sum_{x\in \mathcal{X}} \left \|\textbf{T} \cdot \textbf{S} \cdot \Phi(x)-x \right \|^2,
\end{equation}
where $\mathcal{X}$ represents the sampled points. Note that the scaling matrix $\textbf{S}$ is anisotropic, so it allows more flexible and accurate size estimation compared to NOCS whose scaling matrix is isotropous.

\begin{figure*}[t] \centering
	\begin{overpic}[width=1.0\linewidth,tics=10]{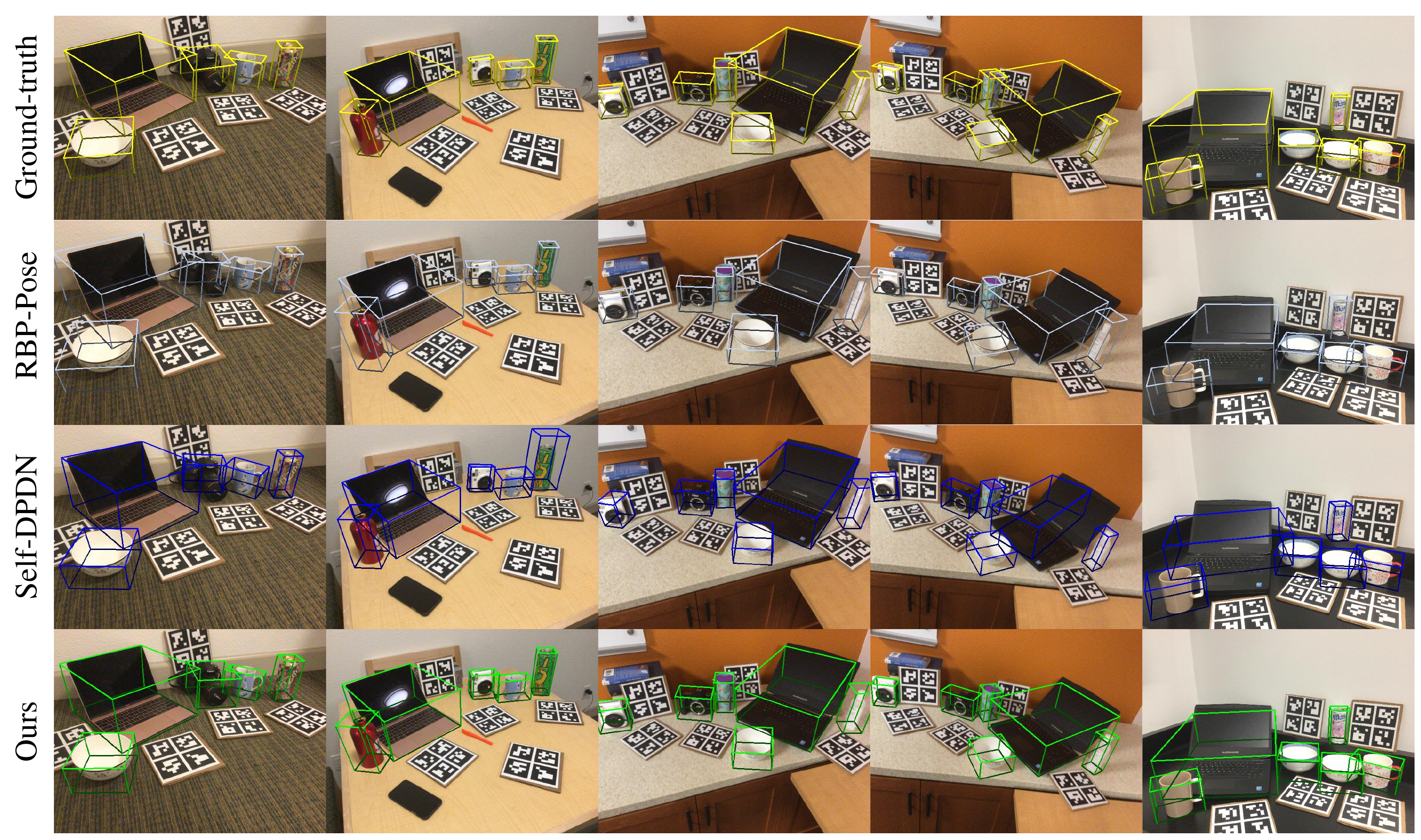}
   \end{overpic}
   \caption{Visual comparison of the estimated pose by our method, RBP-Pose~\cite{zhang2022rbp}, and Self-DPDN~\cite{lin2022category}.}
   \label{fig:qual}
\end{figure*} 

\subsection{Implementation details}
We detected $32$ keypoints on each object.
The multi-scale coordinate-based attention network takes $1,024$ points as input.
The number of classification bins is $128$.
The network is optimized by a ranger optimizer, with batch size $16$ and learning rate $0.001$. The learning rate is annealed at $50\%$ of the training phase using a cosine schedule. We train individual models for each category respectively.
In the surface-independent sampling, we randomly sample points in a sphere with the center being the center of input points and its diameter being the diagonal length of the largest shape in the category.
We set $w_\text{SOCS}$ as $1$ and $w_\text{consistency}$ as $0.1$.


\section{Results and Evaluation}
\label{sec:result}

\subsection{Experimental Datasets}
We train and test our method on the NOCS-REAL275~\cite{wang2019normalized} and ModelNet40-partial~\cite{li2021leveraging} datasets.
The NOCS-REAL275 contains $4.3$k training RGB-D images and $2.75$k testing RGB-D images captured from $6$ real-world scenes. The objects belong to six object categories: bottle, bowl, can, camera, laptop, and mug.
The ModelNet40-partial dataset is a synthetic dataset that contains $60$k training depth images and $6$k testing depth images. It contains object categories with large shape variations, such as \emph{airplane}, \emph{chair}, and \emph{sofa}.

\subsection{Evaluation Metrics}
We use standard metrics to evaluate the performance on the two datasets, respectively.
For NOCS-REAL275, we adopt the intersection over union (IoU) with a threshold of $e$, and the average precision of instances for which the error is less than $n^{\circ}$ for rotation and $m$ for translation.
For ModelNet40-partial, we report the rotational error, and the translational error in the form of mean, and median values. We also report the average precision of instances for which the error is less than $5^{\circ}$ for rotation and $5$cm for translation.

\begin{table*}[!ht]
\renewcommand{\arraystretch}{1.3}
\centering
    \caption{Quantitative results on the NOCS-REAL275 dataset.}
\begin{tabular}{cccccccccc}
\hline
Methods & Data type & Data source & IoU50 $\uparrow$& IoU75$\uparrow$
&$5^{\circ}$2cm$\uparrow$ & $5^{\circ}$5cm$\uparrow$ & $10^{\circ}$2cm$\uparrow$ & $10^{\circ}$5cm$\uparrow$ \\
\hline
          NOCS~\cite{wang2019normalized} & RGB & Syn.+Real & 0.78 & 0.30 & 0.07 & 0.10 & 0.14 &  0.25\\
          SGPA~\cite{chen2021sgpa} & RGB-D & Syn.+Real & 0.80 &0.62& 0.36 & 0.40 & 0.61 & 0.71   \\
          Self-DPDN~\cite{lin2022category} & RGB-D & Syn.+Real & \textbf{0.83} & \textbf{0.76}& 0.46 & 0.51 & 0.70 & 0.78   \\
          GPV-Pose~\cite{di2022gpv} & D & Real  & 0.83 & 0.64 &  0.32 &
          0.43 & - & 0.73 \\
          RBP-Pose~\cite{zhang2022rbp} & D & Real & 0.83 & 0.68 & 0.38 & 0.48 & 0.63 & 0.79  \\
          \hline
          Network in~\cite{wang2019normalized} + SOCS est. & RGB & Real  & 0.79 &0.41 &0.11 &0.12 & 0.15 & 0.30    \\
          Our network + NOCS est. & D & Real & 0.82 & 0.73 &0.40 & 0.49 & 0.64 & 0.81   \\
          Ours  & D & Real  & 0.82 & 0.75 & $\textbf{0.49} $ & \textbf{0.56} &    \textbf{0.72} & \textbf{0.82}  \\
\hline
\end{tabular}
\label{tab:nocs_real}
\end{table*}

\begin{table*}[!t]
\renewcommand{\arraystretch}{1.3}
\centering
    \caption{Quantitative results on the ModelNet40-partial dataset.}
\begin{tabular}{cccccc|ccc}
\hline
\multirow{2}{*}{Methods} & \multirow{2}{*}{Data type} &\multirow{2}{*}{Data source} & \multicolumn{3}{c|}{Rotation} & \multicolumn{3}{c}{Translation} \\
\cline{4-9}
& & &Mean($^{\circ}$) $\downarrow$ & Median($^{\circ}$)$\downarrow$ & $5^{\circ}$ $\uparrow$ & Mean()$\downarrow$ &Median()$\downarrow$ & $5^{\circ}$0.05 $\uparrow$ \\
\hline
EPN~\cite{li2021leveraging} & D & Syn. & 32.86 & 23.84 & 0.49 & 0.14 & 0.13 & 0.08 \\
KPConv~\cite{li2021leveraging} & D & Syn. & 37.48 & 30.86 & 0.24 & 0.11 & 0.08 & 0.06 \\
GPV-Pose~\cite{di2022gpv}&  D & Syn. & 30.75 & 30.41 & 0.28 & 0.17 & 0.11 & 0.06 \\
RBP-Pose~\cite{zhang2022rbp} & D  & Syn. & 33.09 & 35.25 & 0.26 & 0.08 & 0.13 & 0.10 \\
\hline
Ours&D  & Syn. & \textbf{22.53} & \textbf{22.81} & \textbf{0.59} &\textbf{ 0.03 }& \textbf{0.07} & \textbf{0.26} \\
 \hline
\end{tabular}
\label{tab:modelnet40}
\end{table*}

\subsection{Performance on NOCS-REAL275}
We first compare our method with the state-of-the-art on the NOCS-REAL275 dataset.
The quantitative results are shown in Table~\ref{tab:nocs_real}.
There are several phenomena we can observe.
First, Self-DPDN~\cite{lin2022category} slightly outperforms our method on metrics of IoU50 and IoU75, showing that their method is better than ours in terms of object detection.
Second, our method outperforms all the baselines on metrics of $5^{\circ}2$cm, $5^{\circ}5$cm, $10^{\circ}2$cm, $10^{\circ}5$cm, demonstrating the effectiveness of our method on pose estimation despite the inferiority on object detection.
In particular, to further study the effectiveness of the proposed SOCS and the proposed network, we replace each of them with NOCS and the network in~\cite{wang2019densefusion} respectively (i.e. the baseline of Network in~\cite{wang2019normalized} + SOCS est. and Our network + NOCS est.), and conduct experiments.
The results show that our full method is better than the two baselines, revealing the necessity of both the SOCS and the proposed network.
We also found our method requires less training time compared to the baseline of Our network + NOCS est., demonstrating SOCS is easy to train compared to NOCS.
The qualitative comparisons to the state-of-the-art are visualized in \Fig{qual}.


\subsection{Performance on ModelNet40-partial}
\begin{figure}[t] \centering
\begin{overpic}[width=1.0\linewidth,tics=10]{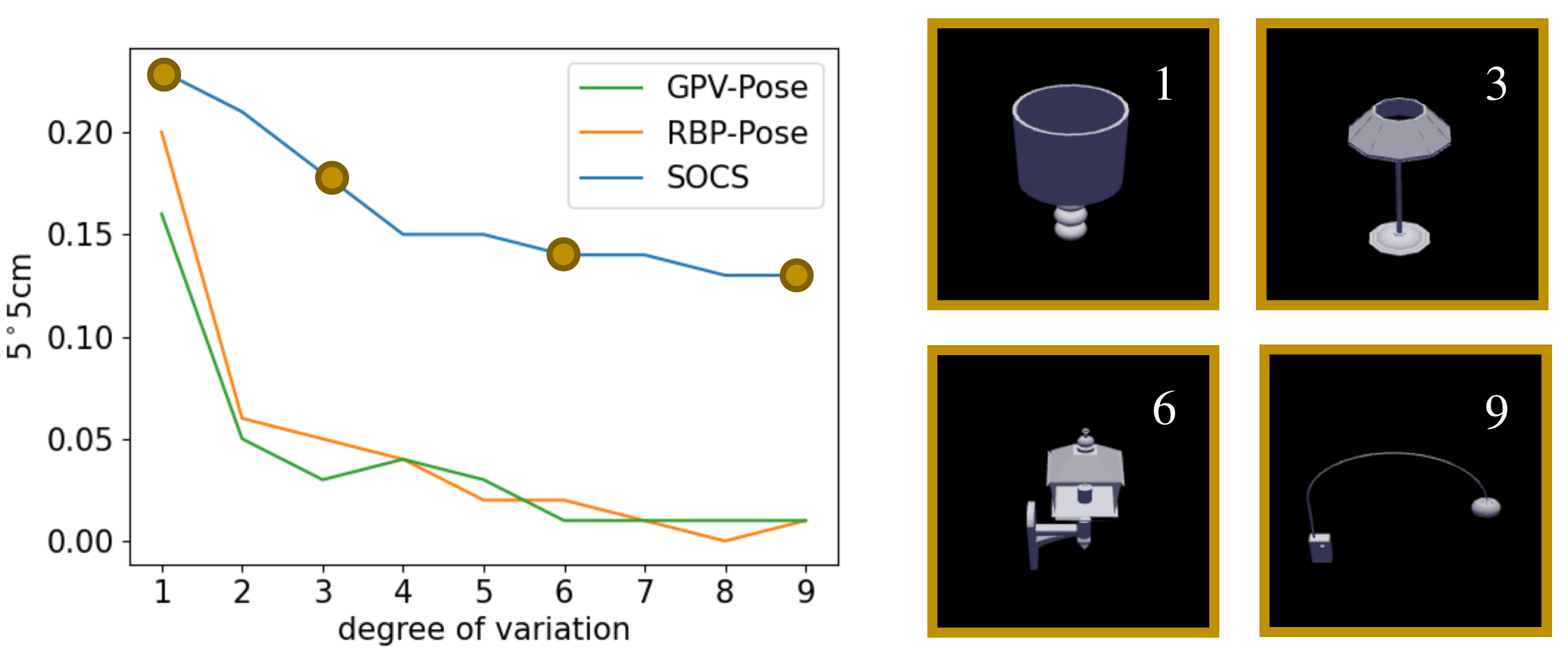}
   \end{overpic}
   \caption{Left: Comparisons on subsets with different degrees of shape variations. We see our method outperforms the baselines on all subsets. Right: Examples of objects instances of different degrees of variations.}
   \label{fig:var_vis}
\end{figure}

To demonstrate the performance of our method under \emph{large shape variations}, we conduct experiments on the ModelNet40-partial dataset. The results are reported in Table~\ref{tab:modelnet40}.
We see that our method outperforms all baselines by a large margin over all the metrics, which suggests that our method is much more effective in handling categories with large shape variations.
Moreover, to quantitatively analyze how our method performs under different shape variations, we conduct an additional experiment.
Specifically, we generate several subsets of the $lamp$ category in the ModelNet40 dataset with different degrees of shape variations.
The degree of shape variations is computed as the average chamfer distance between every shape instance and the categorical mean shape.
The results are visualized in \Fig{var_vis}.
It shows that our method is able to handle object instances with different degrees of shape variations, while the baselines cannot.

\begin{table}[!t]
\renewcommand{\arraystretch}{1.3}
\centering
\caption{Ablation studies of the key components.}
\begin{tabular}{ccccccc}
\hline
  & MP & GP & CL & Sampling & IoU75$\uparrow$ &  $10^{\circ}$2cm$\uparrow$ \\
\hline
$A_1$ &  - & \checkmark  & \checkmark & \texttt{SI} & 0.66& 0.61  \\
$A_2$ &  \checkmark & -  & \checkmark  & \texttt{SI}  & 0.70 & 0.65 \\
    $A_3$ & \checkmark & \checkmark  & - &  \texttt{SI} & 0.72 & 0.67 \\
\hline
$B_1$ & \checkmark &  \checkmark & \checkmark&  \texttt{P}  & 0.67 & 0.62 \\
$B_2$& \checkmark &  \checkmark & \checkmark&  \texttt{SD} & 0.66 & 0.63\\
    \hline
Ours& \checkmark &  \checkmark & \checkmark&  \texttt{SI} & \textbf{0.75} & \textbf{0.72} \\
\hline
\end{tabular}
\label{tab:ablation}
\end{table}

\subsection{Ablation Studies and Parameter Setting}
In Table~\ref{tab:ablation}, we conduct ablation and parameter setting studies to quantify the efficacy of the key components in our method.
In Table~\ref{tab:keypoints} and \ref{tab:bins}, we study the key parameter settings.
All the experiments are conducted on the NOCS-REAL275 dataset.


\paragraph{Network architecture.}
We then study the necessity of crucial network modules, i.e. the multi-block feature propagation (MP), the global position encoding in cross-attention (GP), and the consistency loss function (CL). In the experiment, we remove these crucial modules respectively, retrain the networks, and evaluate the performances.
Note that, in the ablation baseline of MP, we only perform feature propagations at the last block, so it has no multi-block contextual features.
The experiments show that adding any of the modules would lead to a performance improvement, confirming the effectiveness of these network modules.

\vspace{-8pt}

\begin{figure}[t] \centering
\begin{overpic}[width=1.0\linewidth,tics=10]{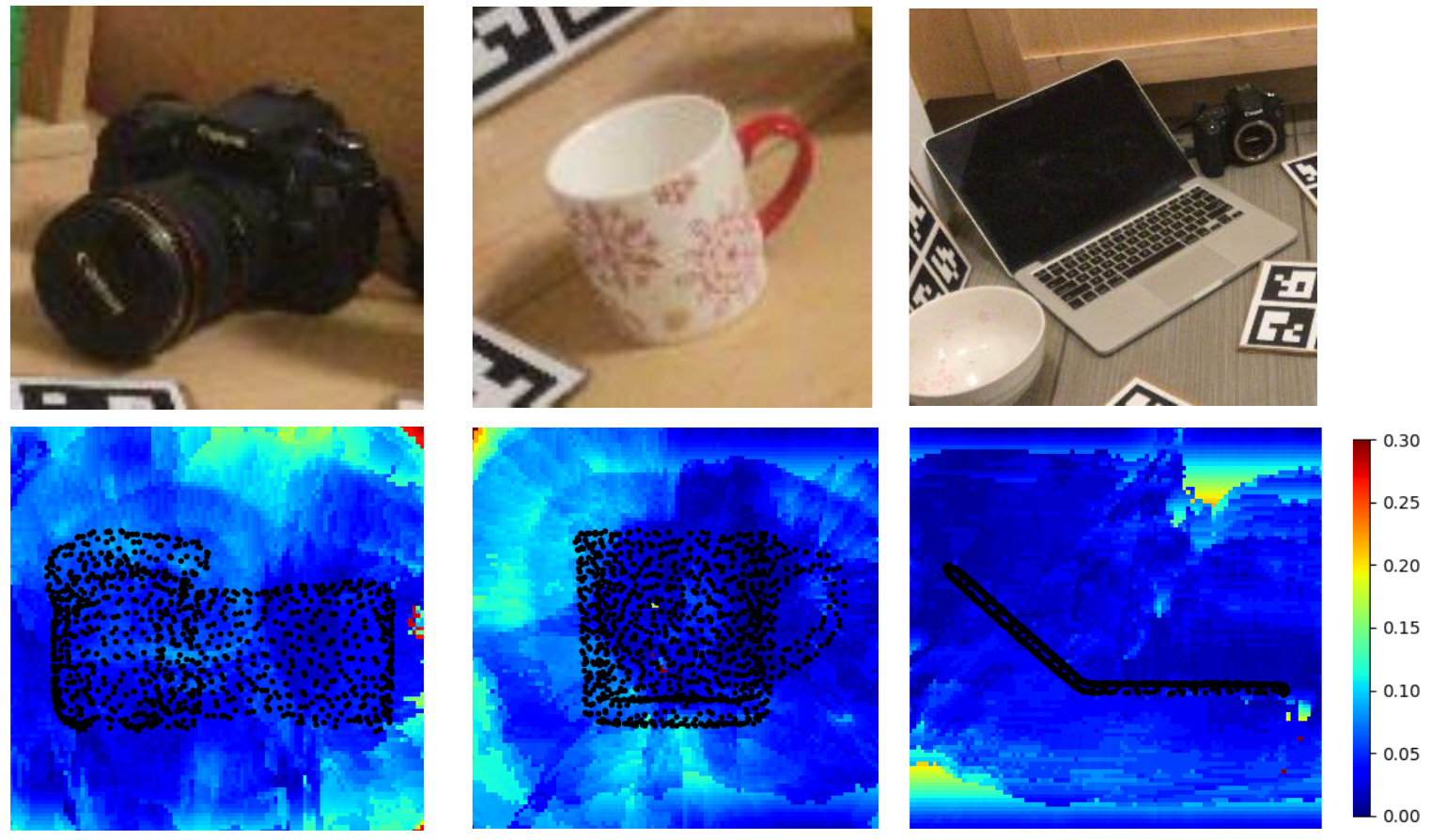}
   \end{overpic}
   \caption{The per-point SOCS estimation error in a cross-section (Bottom row) of the input scene (Top row). The estimation in most of the unseen regions is as accurate as that near the observed surface, showing the necessity of surface-independent sampling and the efficacy of our feature propagation mechanism.}
   \label{fig:error_vis}
\end{figure}

\paragraph{Sampling strategy.}
We have discussed the advantages of the surface-independent sampling (\texttt{SI}) strategy in Section~\ref{sec:method}. Here, we quantitatively compare it with the alternatives of sampling from the input points (\texttt{P}) and surface-dependent sampling (\texttt{SD}). We see that the network trained by the surface-independent sampling strategy outperforms the rest. Moreover, we visualize the per-point SOCS estimation error in a cross-section in \Fig{error_vis}. It is clear that the estimation in most of the unseen regions is as accurate as that near the observed surface, showing the necessity of surface-independent sampling and the efficacy of our feature propagation mechanism.

\begin{table}[!t]
\renewcommand{\arraystretch}{1.3}
\centering
\caption{Effect of different numbers of keypoints. }
\begin{tabular}{ccccc}
\hline
  $\#$keypoints & IoU75$\uparrow$&$5^{\circ}$2cm$\uparrow$ &$5^{\circ}$5cm$\uparrow$  & $10^{\circ}$2cm$\uparrow$ \\
\hline
8 &  0.72 &0.42&0.50 & 0.64  \\
   16 & 0.72 &0.46& 0.54& 0.68 \\
   32 & 0.75 &\textbf{0.49}&\textbf{0.56} & \textbf{0.72}  \\
   64 & \textbf{0.75}  &0.47& \textbf{0.56}& 0.68 \\
        \hline
   32 (ISRP~\cite{chen2020unsupervised}) & 0.71 & 0.42&  0.48& 0.65  \\
\hline
\end{tabular}
\label{tab:keypoints}
\end{table}
\vspace{-8pt}

\begin{table}[!t]
\renewcommand{\arraystretch}{1.3}
\centering
    \caption{Effects of different numbers of classification bins.}
\begin{tabular}{ccccc}
\hline
$\#$bins  & IoU75 $\uparrow$ & $5^{\circ}$2cm$\uparrow$& $5^{\circ}$5cm$\uparrow$
& $10^{\circ}$2cm $\uparrow$ \\
\hline
32& 0.70 & 0.44&0.52 & 0.61 \\
64 & 0.71 & 0.47& 0.55& 0.64 \\
128& \textbf{0.75} & \textbf{0.49}& \textbf{0.56} & \textbf{0.72} \\
256 & 0.73 &\textbf{0.49}& 0.55 & \textbf{0.72}  \\
        \hline
        Regression & 0.69 &0.43&0.50 & 0.66  \\
\hline
\end{tabular}
\label{tab:bins}
\end{table}
\vspace{-8pt}

\paragraph{Number of keypoints.}
The number of keypoints is a crucial parameter that has the potential to influence the effects of SOCS.
We conduct several experiments using different numbers of keypoints to generate the SOCS and retrain our network.
As reported in Table~\ref{tab:keypoints}, we see that using a relatively small number of keypoints would lead to a significant performance decrease. The reason might be that an insufficient number of keypoints would lead to inaccurate dense correspondence between object instances.
We also tried to adopt an alternative keypoints extraction method, i.e. ISRP~\cite{chen2020unsupervised}, instead of Skeleton Merger~\cite{shi2021skeleton}.
Results show the alternative key-point extraction method is also applicable to our method but will lead to inferior performances, implying that the quality of keypoints is crucial to our method.


\paragraph{Number of classification bins.}
We conduct a comparison with baselines that use different numbers of bins in the coordinate classification, as well as a baseline that replaces the classification with regression.
As reported in Table~\ref{tab:bins}, we found that using classification is a better choice compared to using regression.
Besides, the performances reach their peak when the number of classification bins is $128$ or $256$, showing that our method is able to be compatible with a relatively large number of classification bins. This reveals that SOCS indeed simplifies and facilitates the network training.

\subsection{Performance under Occlusion}
Our method is designed to handle moderate inter-object occlusions. In order to verify this, we evaluate our method and compare it to the state-of-the-arts on a subset containing objects with heavy occlusions. To be specific, we select $500$ RGB-D images with instances that have been heavily occluded ($\leq 30\%$ object surface can be observed) from the NOCS-REAL275 dataset. The visualization of the examples in this subset is provided in the supplemental material. In Table~\ref{tab:occlusion}, we see that our method outperforms the state-of-the-arts by a large margin, verifying the ability of our method in terms of handling occlusions.

\begin{table}[!t]
\renewcommand{\arraystretch}{1.3}
\centering
\caption{Comparisons under heavy occlusion.}
\begin{tabular}{ccccccc}
\hline
Method  & IoU75 $\uparrow$ &$5^{\circ}$2cm$\uparrow$&$5^{\circ}$5cm$\uparrow$& $10^{\circ}$2cm $\uparrow$ \\
\hline
RBP-Pose~\cite{zhang2022rbp} & 0.60  &0.29& 0.39& 0.42 \\
Self-DPDN~\cite{lin2022category} & 0.61&0.29&0.40 & 0.47  \\
Ours  & \textbf{0.66} &\textbf{0.38}& \textbf{0.51}&  \textbf{0.55}  \\
\hline
\end{tabular}
\label{tab:occlusion}
\end{table}
\vspace{-8pt}

\section{Conclusion}
\label{sec:conclusion}

We have presented a method for accurate and robust category-level 6D pose and size estimation based on the novel Semantically-aware Object Coordinate Space (SOCS). Since SOCS is built by non-rigidly aligning objects based on
semantically meaningful correspondences, it is semantically coherent and leads to accurate pose and size estimation under large shape variations. In the future, we would like to investigate techniques to boost the performance of our method on objects with complex symmetry. We would also like to explore applying SOCS to the problem of category-level pose estimation of articulated objects.

{\small
\bibliographystyle{ieee_fullname}
\bibliography{egbib}
}

\end{document}